\documentclass[10pt, a4paper]{article}

\usepackage{lrec-coling2024} 
\usepackage{longtable}
\usepackage{booktabs}
\usepackage{svg}
\usepackage{multirow}
\usepackage{float}
\usepackage[T2A,T1]{fontenc}

\usepackage[labelfont=bf]{caption}
\usepackage[framemethod=TikZ]{mdframed}
\usepackage{amsmath,amssymb}
\usepackage{mdframed}
\usepackage{arydshln}
\usepackage{breakurl}
\hypersetup{
           breaklinks=true,   
           colorlinks=true,   
           pdfusetitle=true,  
}

\newcommand{\coedit}{\textsc{CoEdIT}}
\newcommand{\coeditsp}{\textsc{CoEdIT}\space}
\newcommand{\medit}{\textsc{mEdIT}}
\newcommand{\meditsp}{\textsc{mEdIT}\space}
\newcommand{\method}{\textsc{Spivavtor}}
\newcommand{\methodsp}{\textsc{Spivavtor}\space}

\title{\method: An Instruction Tuned Ukrainian Text Editing Model}

\name{Aman Saini, Artem Chernodub, Vipul Raheja, Vivek Kulkarni} 

\address{Grammarly \\
         firstname.lastname@grammarly.com\\
}

\abstract{ 
We introduce \method, a dataset, and instruction-tuned models for text editing focused on the Ukrainian language. \methodsp is the Ukrainian-focused adaptation of the English-only \coedit ~\cite{raheja-etal-2023-coedit} model. 
Similar to \coedit, \methodsp performs text editing tasks by following instructions in Ukrainian like {\fontencoding{T2A}\selectfont``Виправте граматику в цьому реченні''} and {\fontencoding{T2A}\selectfont``Спростіть це речення''} which translate to ``Correct the grammar in this sentence'' and ``Simplify this sentence'' respectively. This paper describes the details of the \method-Instruct dataset and \methodsp models. We evaluate \methodsp on a variety of text editing tasks in Ukrainian, such as Grammatical Error Correction (GEC), Text Simplification, Coherence, and Paraphrasing, and demonstrate its superior performance on all of them. We publicly release our best-performing models and data as resources to the community to advance further research in this space. \\ \newline  }

\begin{document}

\maketitleabstract

\section{Introduction} 
Recently, there has been an increased focus and substantial progress in developing natural language processing (NLP) models for the Ukrainian language. These include the development of corpora like the Ukrainian Brown Corpus ~\cite{starko-rysin-2023-creating}, toolkits like NLP-UK\footnote{\url{https://github.com/brown-uk/nlp_uk}}, as well as models for word-embeddings, part-of-speech tagging, named entity recognition\footnote{\url{https://huggingface.co/lang-uk}}, machine translation\footnote{\url{https://github.com/Helsinki-NLP/UkrainianLT}}, and pre-trained language models. 

However, many of the aforementioned models are task-specific and do not leverage recent advances in large-scale language models and in-context learning. In particular, Large Language Models (LLMs) have demonstrated a quantum leap in generating fluent and coherent text and have thus been adapted to perform various NLP tasks ~\cite{brown2020language, OpenAI_GPT4_2023, touvron2023llama}. Recently, LLMs have also been leveraged to perform specific text editing tasks such as GEC \cite{wu2023chatgpt, coyne2023analysis, fang2023chatgpt} or Text Simplification~\cite{baez-saggion-2023-lsllama,saggion-etal-2022-findings}. In addition to focusing on singular tasks like the above, some works have attempted to leverage LLMs and instruction fine-tuning to support multi-task text editing \cite{schick2023peer, raheja-etal-2023-coedit, laban2023beyond}. 

\begin{figure}[t!]
\begin{mdframed}
\small
\textbf{Grammatical Error Correction}   \hfill \break
\texttt{\textcolor[HTML]{005AB5}{\textbf{{\fontencoding{T2A}\selectfont Виправте граматику в цьому реченні: А ти, батюшка, стало бути, тут в сторожі?}}}}    \hfill \break
\texttt{\textcolor[HTML]{DC3220}{{\fontencoding{T2A}\selectfont А ти, батюшко, стало бути, тут у сторожі?}}}\hfill \break 
\break
\textbf{Coherence}   \hfill \break
\texttt{\textcolor[HTML]{005AB5}{\textbf{{\fontencoding{T2A}\selectfont Покращіть зв'язність тексту: Тоді він відвідав острови Греції. Хвороба змусила його повернутися до Риму.}}}}    \hfill \break
\texttt{\textcolor[HTML]{DC3220}{{\fontencoding{T2A}\selectfont Потім він відвідав острови Греції, але хвороба змусила його повернутися до Риму.}}}\hfill \break 
\break
\textbf{Text Simplification}   \hfill \break
\texttt{\textcolor[HTML]{005AB5}{\textbf{{\fontencoding{T2A}\selectfont Спростіть речення: вода спочатку холодна, а потім волога.}}}}    \hfill \break
\texttt{\textcolor[HTML]{DC3220}{{\fontencoding{T2A}\selectfont Вода і холодна, і мокра.}}}\hfill \break 
\break
\textbf{Paraphrasing}   \hfill \break
\texttt{\textcolor[HTML]{005AB5}{\textbf{{\fontencoding{T2A}\selectfont Перефразуйте речення: Наскільки я можу судити, обидва ще живі.}}}}    \hfill \break
\texttt{\textcolor[HTML]{DC3220}{{\fontencoding{T2A}\selectfont Наскільки я розумію, вони обидва ще живі.}}}\hfill
\end{mdframed}
\caption{Example input (blue) and output (red) of the text editing tasks that \methodsp can perform. The corresponding English translations can be found in Appendix A, Table 7.}
\label{fig:main_examples}
\end{figure}

There has also been extensive work on leveraging these advances to develop corresponding LLMs focused on the Ukrainian language, the most notable being UAlpaca\footnote{\url{https://huggingface.co/robinhad/ualpaca-7b-llama}}, which builds a Ukrainian counterpart of the popular general-purpose instruction-tuned model -- Alpaca ~\cite{alpaca}. Concurrently and similarly, some research has focused on building and characterizing the capabilities of multi-lingual LLMs which are trained on massively multi-lingual data during the pre-training and instruction-tuning phases~\cite{muennighoff-etal-2023-crosslingual, workshop2023bloom, xue-etal-2021-mt5, li2023bactrianx, wei2023polylm, ustun2024aya}. While these models support instructions in Ukrainian, they do not focus on high-quality text editing tasks but on general-purpose instructions instead, such as sentiment detection, question answering, text generation, etc. However, as noted by ~\citet{raheja2024medit}, such generic instruction-tuned models are not particularly well-suited for nuanced text editing tasks without further task-specific fine-tuning. This highlights the need for an instruction-tuned model for Ukrainian that is optimized for text editing, which this paper addresses by building \method\footnote{\method\ means ``co-author'' in Ukrainian.}. 

\method\ can follow instructions for complex text editing tasks like GEC, Text Simplification, Coherence, and Paraphrasing (Figure \ref{fig:main_examples}). A significant challenge to building an instruction-tuned model for Ukrainian optimized for text editing has been the limited availability of text editing datasets in Ukrainian. In this work, we address this challenge by adapting existing text editing datasets from Ukrainian and English and converting them to ''instruction-following'' datasets (similar to \coeditsp and \medit). We then show how these newly constructed datasets can be used to build state-of-the-art text editing models for Ukrainian. Finally, through comprehensive evaluations, we empirically reveal critical insights on how the performance on Ukrainian text editing tasks is affected by various choices like model architecture, model scale, and training data mixtures. All our models and data are publicly available as resources for the community\footnote{\url{https://huggingface.co/collections/\\grammarly/spivavtor-660744ab14fdf5e925592dc7}}.

\section{Related Work}
Prior work falls into two major categories: (a) Ukrainian-NLP Models and (b) Multi-lingual LLMs. We discuss each of these below.

\paragraph{Large Language Models for Ukrainian}
Several works have focused on building LLMs and resources for Ukrainian. These mainly consist of manually curated Ukrainian language datasets and corpora like ~\citet{starko-rysin-2023-creating} for Part of Speech, ~\citet{syvokon-etal-2023-ua} for Grammatical Error Correction (GEC), NER-UK for Named Entity Recognition\footnote{\url{https://github.com/lang-uk/ner-uk}}, UA-SQUAD for Question Answering ~\citet{ua_datasets_2021}. Some Ukrainian datasets are also derived from large multi-lingual datasets filtered for the Ukrainian language data (for e.g., Ukrainian Tweet Corpus\footnote{\url{https://github.com/saganoren/ukr-twi-corpus}}). In addition to these datasets, custom models have also been built for the above tasks, a list of which is curated here\footnote{\href{https://github.com/osyvokon/awesome-ukrainian-nlp}{\texttt{https://github.com/osyvokon/awesome-ukrainian-\\
nlp}}}. A notable such model aimed at general instruction following in Ukrainian is the UAlpaca model, which was obtained by further fine-tuning LLaMA on Ukrainian translations of the Alpaca \cite{alpaca} dataset.

\paragraph{Text Editing via Instruction Tuning}
There exists extensive prior literature leveraging instruction-tuned LLMs for various text editing tasks in both monolingual and multi-lingual settings. More recently, \citet{schick2023peer}, \citet{raheja-etal-2023-coedit}, and \citet{laban2023beyond} have focused on general-purpose text editing using instruction-tuned LLMs for English.
However, all of these prior approaches have been limited in monolingual settings because they focus only on English.

Text editing capabilities in the Ukrainian language have been developed only in multi-lingual settings, where most works have proposed task-specific multi-lingual models. These works have developed models for text editing tasks like GEC  (\citet{rothe-etal-2021-simple}; \citet{sun2022unified}), paraphrasing \cite{chowdhury2022novelty}, formality style transfer \cite{briakou-etal-2021-ola}, and text simplification (\citet{mallinson-etal-2020-zero}; \citet{martin-etal-2022-muss}; \citet{ryan-etal-2023-revisiting}). However, they are similarly limited due to their singular focus on specific text editing tasks rather than high-quality, general-purpose text editing.

There exists an even more extensive literature on general-purpose multi-lingual LLMs (many of which also include support for Ukrainian ~\cite{ustun2024aya, li2023bactrianx}), these models generally aim for massive multi-language support and are not optimized explicitly for Ukrainian or text editing.  A comprehensive review of multi-lingual LLMs is out of the scope of this paper.

Finally, our work is closest to the recently proposed \meditsp  \cite{raheja2024medit}, which developed a multi-lingual extension to \coeditsp with support for a similar set of tasks for six languages, but is limited in our context as it is not focused on Ukrainian as one of its core languages. 


\section{\method}
In this section, we describe the construction of \method.  Specifically, we discuss (a) Dataset construction, (b) Model architecture choices, and (c) Model training process.

\subsection{\method-Instruct Dataset}
Similar to prior work ~\cite{raheja-etal-2023-coedit}, we consider four text editing tasks: (a) Fluency/Grammatical Error Correction (GEC), (b) Simplification, (c) Coherence, and (d) Paraphrasing; and construct a unified Ukrainian text editing instruction dataset which we call \method-Instruct. We consider these tasks for two reasons: (a) These tasks are largely representative of the most common text editing tasks, and (b) It is feasible to obtain curated good-quality data for these tasks either in Ukrainian or English. For tasks where Ukrainian data is not readily available, we use the available English datasets to construct their Ukrainian counterpart by translating them into Ukrainian using Google Translate API\footnote{\href{https://cloud.google.com/translate/docs/advanced/translating-text-v3}{\texttt{https://cloud.google.com/translate/\\docs/advanced/translating-text-v3}}}. Due to time constraints, we did not explore other translation services or models. Having outlined the tasks, we now discuss the task-specific datasets we used and our process for constructing \method-Instruct.

\paragraph{GEC} We use the Ukrainian Grammatical Error Correction (UA-GEC) dataset \cite{syvokon-etal-2023-ua} for GEC/Fluency. This dataset contains $33$k pairs of grammatically incorrect and correct sentences in Ukrainian. The original dataset contains train ($31$k) and test ($2$k) splits. However, since we explore different model choices and training hyper-parameters, we further randomly split the train set to create a custom train ($28$k) and validation ($3$k) dataset. 

\paragraph{Simplification} For the Simplification task, we adapt three English datasets: (a) WikiLarge \cite{zhang-lapata-2017-sentence}, and (b) WikiAuto \cite{jiang-etal-2020-neural} for training. For evaluation, we use ASSET \cite{alva-manchego-etal-2020-asset}, and Turk ~\cite{xu2016optimizing} datasets. As mentioned above, we translate all these datasets into Ukrainian using Google Cloud Translation API.

\paragraph{Coherence} For the coherence task, which involves combining two sentences together coherently using edit operations such as inserting discourse connectives, we once again translate an English dataset, given the lack of an equivalent dataset for Ukrainian. In particular, we adapt the DiscoFuse dataset \cite{geva-etal-2019-discofuse} and the Coherence split of \textsc{IteraTeR} \cite{du-etal-2022-understanding-iterative} and translate them to Ukrainian using the Google Cloud Translation API. 

\paragraph{Paraphrasing} We adapt the popular PAWS  \cite{zhang-etal-2019-paws} dataset in English by constructing its Ukrainian counterpart via translation, maintaining their train and test splits. We evaluate paraphrasing on MRPC \cite{dolan-brockett-2005-automatically}, STS \cite{cer-etal-2017-semeval}, and QQP datasets.

\medskip

\noindent The Ukrainian datasets we thus obtain are suitable for training Ukrainian-specific models, but they are not suitable yet for instruction tuning since they do not contain explicit instructions. To overcome this, we prepend task-specific verbalizers that describe the task to be performed as simple instructions to each instance. These task-specific verbalizers were curated by domain experts in Ukrainian. More specifically, for a given task-specific instance, we assign a specific verbalizer by randomly drawing a sample from the task-specific verbalizer set. Table \ref{tab:verbalizers_small} shows a few instruction verbalizers for each task with the full set available in Appendix Table \ref{tab:verbalizers_full}. Similarly, Table \ref{tab:dataset_stats} summarizes the number of training, validation, and test instances, along with the number of distinct instructions per task. Finally, it is to be noted that to ascertain the quality of the Ukrainian translated datasets, a random sample of $100$ instances were chosen for verification by native speakers of Ukrainian and found to be largely satisfactory\footnote{Grossly incorrect translations were corrected manually.}.

\begin{table*}[htb!]
\centering
\begin{tabular}{lcccc}
\toprule
\textbf{Task} & \textbf{\#Train} & \textbf{\#Validation} & \textbf{\#Test} & \textbf{\#Verbalizers} \\ 
\toprule
\textbf{GEC} & 27,929 & 3,103 & 2,682 & 9 \\
\textbf{Simplification} & 11,501 & 1,278 &  533 & 11 \\
\textbf{Coherence} & 9,278 & 1,031 & 551 & 7 \\
\textbf{Paraphrasing} & 14,076 & 1,564 & 6,244 & 13 \\ 
\toprule
\textbf{Total} & 62,784 & 6,976 & 10,010 & 40 \\ 
\bottomrule
\end{tabular}
\caption{Summary statistics of the \method-Instruct dataset.}
\label{tab:dataset_stats}
\end{table*}

\begin{table*}[htb!]
\centering
\begin{tabular}{l|l|l}
\toprule
\textbf{Task} & \textbf{Verbalizers} & \textbf{English translation}\\
\midrule
\textbf{GEC} & \begin{tabular}[c]{@{}l@{}}{\fontencoding{T2A}\selectfont``Виправте граматику в цьому реченні:''} \\ {\fontencoding{T2A}\selectfont``Зробіть речення граматичним:''} \\ {\fontencoding{T2A}\selectfont``Удосконаліть граматику цього тексту:''}\end{tabular} 
& \begin{tabular}[c]{@{}l@{}}``Correct the grammar in this sentence:'' \\ ``Make the sentences grammatical:'' \\ 
``Improve the grammar of this text:''\end{tabular} 
\\ 
\midrule
\textbf{Simplification} & \begin{tabular}[c]{@{}l@{}}{\fontencoding{T2A}\selectfont``Спростіть речення:''}\\
{\fontencoding{T2A}\selectfont``Зробіть речення простим:''}\\
{\fontencoding{T2A}\selectfont``Зробіть цей текст легше для розуміння:''}\end{tabular} 
& \begin{tabular}[c]{@{}l@{}}``Simplify the sentences:''\\
``Make the sentence simple:''\\
``Make this text easier to understand:''\end{tabular}
\\ 
\midrule 
\textbf{Coherence} & \begin{tabular}[c]{@{}l@{}}{\fontencoding{T2A}\selectfont``Виправте зв'язність в реченні:''}\\
{\fontencoding{T2A}\selectfont``Покращіть зв'язність тексту:''} \\
{\fontencoding{T2A}\selectfont``Зробіть текст більш зв'язним:''}\end{tabular} 
& \begin{tabular}[c]{@{}l@{}}``Correct the coherence in the sentence:''\\
``Improve text coherence:'' \\
``Make the text more coherent:''
\end{tabular}
\\ 
\midrule
\textbf{Paraphrasing} & \begin{tabular}[c]{@{}l@{}}{\fontencoding{T2A}\selectfont``Перефразуйте речення:''}\\
{\fontencoding{T2A}\selectfont``Перефразуйте цей текст:''} \\
{\fontencoding{T2A}\selectfont``Напишіть перефраз для речення:''} \end{tabular} 
& \begin{tabular}[c]{@{}l@{}}``Rephrase the sentence:''\\
``Paraphrase this text:'' \\
``Write a paraphrase for the sentence:'' \end{tabular}
\\ 
\bottomrule
\end{tabular}
\caption{A subset of verbalizers for each task used as instructions in the \method-Instruct dataset (see Appendix Table \ref{tab:verbalizers_full} for full set of instructions). }
\label{tab:verbalizers_small}
\end{table*}

\subsection{Models}
To train \method, we consider two kinds of transformer-based LLM architectures -- Encoder-Decoder as well as the Decoder-only architecture.
Both architectures have been shown to be generally effective in prior work ~\cite{xue-etal-2021-mt5,ustun2024aya} although the Decoder-only models tend to be more popular recently with the release of models like ChatGPT and GPT4 ~\cite{OpenAI_GPT4_2023}. Thus, in the realm of Ukrainian text editing, we empirically explore the effect of both of these model architectures on task performance. We also explore the effect of different model sizes considering relatively smaller models with 1B parameters as well as larger models with upto 13B parameters. 

\subsubsection{Encoder-Decoder Models}
\paragraph{mT5} \cite{xue-etal-2021-mt5} is a multi-lingual variant of T5 \cite{2020t5}, trained on the mC4 dataset \footnote{\url{https://huggingface.co/datasets/mc4}}, a multi-lingual variant of the C4 dataset extended to 101 languages. We experiment with two variants of mT5 -- \textsc{large} (1.2B) and \textsc{xxl} (13B). 

\paragraph{mT0} \cite{muennighoff-etal-2023-crosslingual} is a family of multi-lingual Encoder-Decoder models capable of following human instructions in dozens of languages. We use the mt0-\textsc{large} (1.2B) model. The mT0 models are constructed by fine-tuning mT5 models on the xP3 cross-lingual task mixture dataset, which consists of multi-lingual datasets with English prompts. As a result, mT0 models are better suited for following English prompts. We also use the mt0-\textsc{xxl}-mt variant, which is fine-tuned on the xP3mt dataset and is better suited for prompting in non-English.


\paragraph{Aya 101} \cite{ustun2024aya} is a massively multi-lingual generative language model that follows instructions in 101 languages of which over 50\% are considered low-resourced. Aya outperforms mT0 and BLOOMZ \cite{muennighoff2022crosslingual} on the majority of tasks while covering double the number of languages. The model has 13B parameters and the same architecture as the mt5-\textsc{xxl}  model. 

\subsubsection{Decoder-only LLMs}
\paragraph{Bactrian-X} \cite{li2023bactrianx} is a collection of lightweight adapters for LLaMA (7B and 13B) \cite{touvron2023llama} and BLOOM (7B) \cite{workshop2023bloom} on the Bactrian-X dataset, which is a multi-lingual parallel dataset containing 3.4 million instruction–response pairs across 52 languages. We use the bactrian-x-llama-7b-merged variant.

\paragraph{Mistral} \cite{jiang2023mistral} is a family of large language models. We use the Mistral-7B-Instruct-v0.2 variant which is an instruction fine-tuned version of the Mistral-7B-v0.2 model.

\paragraph{Llama2 Chat Models} We also consider full-parameter fine-tuning of the Llama2 7B and 13B chat models. While the aforementioned Bactrian-X models also derive from the LLaMA models, they use parameter-efficient fine-tuning (PEFT), specifically, low-rank adaptation (LoRA) \cite{hu2022lora}, thus, significantly reducing the number of trainable parameters during fine-tuning. Thus, in contrast to Bactrian-X models, we consider full-parameter fine-tuning of Llama-2 Chat models as well. We use the Llama-2-7b-chat-hf and Llama-2-13b-chat-hf variants.

\subsection{Training}
We use \method-Instruct dataset to perform instruction-tuning on both styles of models described above. We train all models using Deepspeed \cite{10.1145/3394486.3406703} on 8xA100 GPU instances with AdamW optimizer, a per-device batch size of 8, and a learning rate of 5e-5. For Decoder-only models, the maximum sequence length is set to 512 tokens, whereas for Encoder-Decoder models, the maximum sequence length is set to 256 tokens for both source and target. The best-performing checkpoints were chosen based on
the validation loss. 

\subsection{Inference}
For Inference, we mostly use default generation parameters for temperature, beam size as specified in the corresponding model with the exception of max output length, which is set to the max sequence length used while training the model. 
To avoid repeated generation with Decoder-only models, we used the model-specific EOS tag to end decoding.

\section{Evaluation}
\paragraph{Metrics} We evaluate all models on the task-specific test splits of the \method-Instruct dataset. As in prior work, we report the standard evaluation metrics used for each task. In particular, we report the $F_{0.5}$ Correction score for GEC calculated using ERRANT \cite{bryant-etal-2017-automatic} weighing precision twice as much as recall. Following prior work by \citet{ryan-etal-2023-revisiting, raheja-etal-2023-coedit} we report SARI \cite{xu-etal-2016-optimizing} for Simplification as well as Coherence. For Paraphrasing, we report BLEU \cite{papineni-etal-2002-bleu}. In order to capture the overlap with source as well as reference, we report both reference-free BLEU  (also called Self-BLEU in \citealt{zhu2018texygen}) and reference-based BLEU, since they collectively provide additional signal on paraphrasing quality than either one of them alone (see ~\citet{shen2022evaluation} and Section~\ref{sec:limitations}).  

\paragraph{Baselines} We evaluate our \method\ models against strong instruction tuned baseline models. In addition to the corresponding base models (i.e. not fine-tuned on \method-Instruct dataset), we also evaluate against the following:  

\begin{itemize}
\item \textbf{Copy}: The Copy baseline, which just copies the input sentence, is a surprisingly trivial but hard-to-beat baseline. 

\item \textbf{UAlpaca}: To ascertain the effect of task-specific instruction fine-tuning in contrast to large-scale diverse instruction fine-tuning, we consider the UAlpaca model in a zero-shot setting. UAlpaca is a LLaMA 7B model trained on Ukrainian translations of $52$K diverse and generic instructions of the Alpaca dataset~\cite{alpaca}. For prompting UAlpaca, we used the recommended prompt format that it was fine-tuned on and replaced the instruction placeholder with the assigned verbalizer. 

\item \textbf{GPT4} Noting the widespread popularity of GPT4 \cite{OpenAI_GPT4_2023} and a general notion that GPT4  generally obtains very strong performance on many NLP tasks, we also consider this as a baseline (in the zero-shot setting) where we prompt it with a verbalizer and the input text. In particular, we use gpt-4-0613 model with a context window of 8192 tokens and a training data cutoff of Sep 2021. To give GPT4 the best shot at success and to account for prompt sensitivity, we evaluate GPT4 on the chosen task with all possible verbalizers in our set and report the score corresponding to the best verbalizer. If there is no response received from the API due to content filtration policies, we consider the input unchanged for evaluation purposes. 

\item \textbf{GPT-3.5-Turbo} We also compare against the more cost effective GPT-3.5-Turbo model, widely known as ChatGPT. In particular, we use gpt-3.5-turbo version 0301.

\end{itemize}

\begin{table*}[htb!]
\begin{tabular}{lcccccc}
\toprule
\textbf{Model} & \textbf{Type} & \textbf{Size} & \textbf{GEC} & \textbf{Simplification} & \textbf{Coherence} & \textbf{Paraphrasing} \\ 
\toprule
Copy & - & - & 0 & 21.98 & 26.89 & 
100/31.4 \\ 
\toprule
\textsc{Bactrian-X-7b} & D & 7B & 0.65 & 36.76 & 40.37 & 21.86/8.13 \\
\textsc{UAlpaca-7b} & D & 7B & 0.57 & 35.17 & 32.64 & 13.26/4.95 \\
\textsc{Mistral-7b} & D & 7B & 0.3 & 38.96 & 32.41 & 9.30/3.79 \\ 
\textsc{mt0-large} & ED & 1.2B & 0.21 & 29.56 & 22.14 & 6.70/2.68 \\
\textsc{aya-101} & ED & 13B & 21.98 & 35.59 & 38.30 & 42.68/15.53 \\
\toprule
\textsc{GPT-3.5-Turbo} & D & - & 1.17 & 40.18 & 44.93 & 26.60/12.51 \\
\textsc{GPT4} & D & - & 27.18 & 40.08 & 43.44 & 23.23/11.7 \\ 
\toprule
\textsc{\method-Bactrian-X-7b} & D & 7B & 55.73 & 36.90 & 47.80 & 65.31/23.65 \\
\textsc{\method-Mistral-7b} & D & 7B & 51.54 & 34.55 & 44.12 & 76.56/25.33 \\
\textsc{\method-Llama2-7b} & D & 7B & 55.88 & 36.94 & 48.73 & 48.97/18.9 \\
\textsc{\method-Llama2-13b} & D & 13B & 56.48 & 36.98 & 48.55 & 57.31/21.35 \\
\textsc{\method-mt5-large} & ED & 1.2B & 61.83 & 36.40 & 48.27 & 77.31/26.68 \\
\textsc{\method-mt0-large} & ED & 1.2B & 61.44 & 36.16 & 48.28 & 77.83/26.73 \\
\textsc{\method-mt5-xxl} & ED & 13B & 63.00 & 37.84 & 48.97 & 72.42/25.64 \\
\textbf{\textsc{\method-mt0-xxl-mt}} & ED & 13B & 64.55 & \textbf{38.44} & \textbf{49.48} & 68.63/25.07 \\
\textbf{\textsc{\method-aya-101}} & ED & 13B & \textbf{64.57} & 37.87 & 48.51 & 73.28/26.17 \\
\bottomrule
\end{tabular}
\caption{Comparison of \methodsp models against various baselines including Copy (target=source), Decoder-only(D) and Encoder-Decoder(ED) models when evaluated in a zero-shot setting. 
For GEC, we report $\mathbf{F_{0.5}}$ \textbf{Correction}. For Simplification and Coherence, we report \textbf{SARI}. For Paraphrasing, we report \textbf{ref-free/ref-based BLEU} where ref-free is the reference-free BLEU and ref-based is the reference-based BLEU to capture the overlap with both source and reference. All scores have been scaled to lie between 0 and 100. Note that all \methodsp models outperform baseline models.}
\label{tab:summary_results}
\end{table*}

\begin{table*}[htb!]
\centering
\begin{tabular}{lllll}
\toprule
\textbf{Held-Out Task} & \textbf{GEC} & \textbf{Simplification} & \textbf{Coherence} & \textbf{Paraphrasing} \\ 
\midrule
GEC & \textbf{18.47} & 37.41 & 52.11 & 71.44/26.14 \\
Simplification & 64.95 & \textbf{32.84} & 48.96 & 68.39/25.01 \\
Coherence & 62.57 & 36.79 & \textbf{39.48} & 72.86/25.81 \\
Paraphrasing & 64.25 & 36.86 & 51.84 & \textbf{74.61/25.90} \\
\midrule
\end{tabular}
\caption{Performance of the \method-aya-101 model on all tasks when one task is ablated. We report the same metrics as in Table \ref{tab:summary_results}. The bolded numbers represent the zero-shot performance of the model when not trained on that particular task.}
\label{tab:held-out}
\end{table*}

\subsection{Quantitative Results}
In this section, we describe our main results and discuss findings from ablation studies to gain insights into the factors driving model performance. 

\paragraph{Main Results} Table \ref{tab:summary_results} shows the performance of various models on all tasks in consideration. It presents aggregated scores for all tasks across different datasets. The dataset-specific scores for all relevant tasks are present in Appendix A, Table 8.  Based on these results, we can make the following observations:
\begin{enumerate}
\item \textbf{\methodsp generally performs significantly better over baselines}. Comparing the performance metrics for \methodsp models to their baseline counterparts, we generally observe that \methodsp significantly outperforms baseline models (including GPT4), with Simplification being the only exception where performance is at par. This result suggests the effectiveness of domain-specific instruction tuning for superior performance on specific tasks. 

\item \textbf{Domain-specific Instruction tuning outperforms instruction tuning on a large set of generic instructions.} 
Given the effectiveness of instruction tuning and in-context learning, a natural question arises: For text editing with instructions, is it sufficient to instruction-tune a model with a very large set of diverse instructions that are not necessarily related to text editing? We can answer this question empirically by comparing the performance of \methodsp models (that are instruction tuned on text editing instructions) with UAlpaca -- a model that is instruction tuned on $52$K diverse instructions. From Table \ref{tab:summary_results}, we observe that UAlpaca has significantly lower performance compared to its equivalent \methodsp model (\method-Llama2-7B). It may not be sufficient to instruction-tune models on just a large set of diverse instructions, and there is significant value to instruction tuning on domain-specific instructions, an observation that reaffirms findings in prior work by ~\citet{raheja-etal-2023-coedit}.  

\item \textbf{Encoder-Decoder models outperform Decoder-only models.} Given the extensive popularity of LLMs, there has been a significant surge in the availability of LLMs. While some LLMs use an Encoder-Decoder architecture ~\cite{xue-etal-2021-mt5,ustun2024aya}, some others use a Decoder-only style ~\cite{OpenAI_GPT4_2023, alpaca,touvron2023llama}. Yet, it is not clear if one architecture offers consistently superior performance over the other and on what tasks one might prefer a specific architecture. We trained both styles of models on the \method-Instruct dataset to evaluate the results empirically. Our results indicate that Encoder-Decoder models generally outperform Decoder-only models when fine-tuned on domain-specific instructions. More specifically, note that all \methodsp Encoder-Decoder models outperform \methodsp Decoder-only models on average.

\item \textbf{Larger models outperform smaller ones.} Our results also suggest that, generally, larger models tend to perform better than smaller ones - both across baselines and across \methodsp models within an architecture family. This finding further reinforces the effectiveness of model scaling on task performance.
\end{enumerate}

\paragraph{Task Ablation} In this setting, we hold out specific tasks in a controlled manner to evaluate one of the \methodsp models (\method-aya-101), to see how it might generalize to unseen text editing tasks. More specifically, in each turn, we hold out one of the tasks, train on the remaining set, and report task performance on all tasks. The results of this ablation study are shown in Table \ref{tab:held-out} and clearly demonstrate the usefulness of instruction tuning on all tasks. The model performs significantly better when trained on task-specific data as compared to the zero-shot setting.

\begin{table*}[htbp!]
\small
\centering
\begin{tabular}{p{0.45\textwidth} p{0.45\textwidth}}
\toprule
\textbf{\textcolor[HTML]{005AB5}{GEC Input}\ }
 $\blacktriangleright$ {\fontencoding{T2A}\selectfont Виправте граматику в цьому реченнi: Дякую за інформацію! ми з Надією саме вийшли з дому} &  \textbf{\textcolor[HTML]{005AB5}{GEC Input}\ } $\blacktriangleright$ {\fontencoding{T2A}\selectfont Correct the grammar in this sentence: Thanks for the information! we with Nadia just left the house}\\
  
 \textbf{\textcolor[HTML]{DC3220}{Output}\ } $\blacktriangleright$ {\fontencoding{T2A}\selectfont Дякую за інформацію! Ми з Надією саме вийшли з дому.} &  \textbf{\textcolor[HTML]{DC3220}{Output}\ } $\blacktriangleright$ {\fontencoding{T2A}\selectfont Thanks for the info! Nadia and I just left the house.} 
\\
\addlinespace
\cdashline{1-2}[1pt/1pt]
\addlinespace


\textbf{\textcolor[HTML]{005AB5}{Simplification Input}\ }
 $\blacktriangleright$ {\fontencoding{T2A}\selectfont Спростiть речення: Там він помер через шість тижнів, 13 січня 888 року.} & \textbf{\textcolor[HTML]{005AB5}{Simplification Input}\ }
 $\blacktriangleright$ {\fontencoding{T2A}\selectfont Simplify the sentence: There he died six weeks later, on January 13, 888.}  \\
\textbf{\textcolor[HTML]{DC3220}{Output}\ } $\blacktriangleright$ {\fontencoding{T2A}\selectfont Там він помер 13 січня 888 року.} & \textbf{\textcolor[HTML]{DC3220}{Output}\ } $\blacktriangleright$ {\fontencoding{T2A}\selectfont There he died on January 13, 888.}  \\
\addlinespace
\cdashline{1-2}[1pt/1pt]
\addlinespace


\textbf{\textcolor[HTML]{005AB5}{Coherence Input}\ }
 $\blacktriangleright$ {\fontencoding{T2A}\selectfont Виправте зв’язнiсть в реченнi: Лінч досі відмовляється розповідати про сумнозвісну травневу дорожньо-транспортну пригоду, коли він збив жінку-пішохода в районі нічного клубу Баффало та відлетів. Той факт, що Лінч взагалі говорив, заслуговує на увагу в цих краях.} & \textbf{\textcolor[HTML]{005AB5}{Coherence Input}\ }  $\blacktriangleright$ {\fontencoding{T2A}\selectfont  Correct the coherence in the sentence: Lynch still refuses to talk about the infamous May traffic accident in which he struck a female pedestrian in a Buffalo nightclub area and drove away. The fact that Lynch spoke at all deserves attention in this place.}\\

\textbf{\textcolor[HTML]{DC3220}{Output}\ } $\blacktriangleright$  {\fontencoding{T2A}\selectfont Лінч досі відмовляється розповідати про сумнозвісну травневу дорожньо-транспортну пригоду, коли він збив жінку-пішохода в районі нічного клубу Баффало та відлетів. Однак той факт, що він взагалі говорив, заслуговує на увагу в цих краях.} & \textbf{\textcolor[HTML]{DC3220}{Output}\ } $\blacktriangleright$ {\fontencoding{T2A}\selectfont Lynch still refuses to talk about the infamous May traffic accident in which he struck a female pedestrian in a Buffalo nightclub area and drove away. However, the fact that Lynch spoke at all deserves attention in this place.} \\
\addlinespace
\cdashline{1-2}[1pt/1pt]
\addlinespace


\textbf{\textcolor[HTML]{005AB5}{Paraphrasing Input}\ }
 $\blacktriangleright$ {\fontencoding{T2A}\selectfont Перефразуйте речення: Який найкращий комплімент, який ти отримував від будь-кого?}  & \textbf{\textcolor[HTML]{005AB5}{Paraphrasing Input}\ }
 $\blacktriangleright$ {\fontencoding{T2A}\selectfont Paraphrase the sentence: What is the greatest compliment that you ever received from anyone?} \\
 
\textbf{\textcolor[HTML]{DC3220}{Output}\ } $\blacktriangleright$ {\fontencoding{T2A}\selectfont Який найкращий комплімент, який ти коли-небудь отримував?} & \textbf{\textcolor[HTML]{DC3220}{Output}\ } $\blacktriangleright$ {\fontencoding{T2A}\selectfont What is the greatest compliment that you ever received?}\\ 

\bottomrule
\end{tabular}
\caption{Example inputs and outputs from \textsc{\method-aya-101} model for all relevant tasks.}
\label{tab:qualitative_outputs}
\end{table*}

\begin{table*}[htb!]
\small
\centering
\begin{tabular}{p{0.45\textwidth} p{0.45\textwidth}}
\toprule

\textbf{\textcolor[HTML]{005AB5}{GEC Input}\ }
 $\blacktriangleright$ {\fontencoding{T2A}\selectfont Виправте граматичнi помилки в цьому реченнi: В поки що вересень будем повну оплату робити.
} & \textbf{\textcolor[HTML]{005AB5}{GEC Input}\ }
 $\blacktriangleright$ {\fontencoding{T2A}\selectfont Correct the grammatical errors in this sentence: On the meantime in September, will we make the full payment.
}  \\
\textbf{\textcolor[HTML]{DC3220}{Output}\ } $\blacktriangleright$ {\fontencoding{T2A}\selectfont У поки що вересні будем повну оплату робити.} & \textbf{\textcolor[HTML]{DC3220}{Output}\ } $\blacktriangleright$ {\fontencoding{T2A}\selectfont In the meantime in September, will we make the full payment.}\\
\addlinespace
\cdashline{1-2}[1pt/1pt]
\addlinespace


\textbf{\textcolor[HTML]{005AB5}{Simplification Input}\ }
 $\blacktriangleright$ {\fontencoding{T2A}\selectfont Зробiть речення простим: Джидда є головними воротами до Мекки, найсвятішого міста ісламу, яке працездатні мусульмани повинні відвідати принаймні раз у житті.} & \textbf{\textcolor[HTML]{005AB5}{Simplification Input}\ }
 $\blacktriangleright$ {\fontencoding{T2A}\selectfont Make the sentence simple: Jeddah is the main gateway to Mecca, Islam's holiest city, which able-bodied Muslims must visit at least once in their lifetime.}  \\
\textbf{\textcolor[HTML]{DC3220}{Output}\ } $\blacktriangleright$ {\fontencoding{T2A}\selectfont Це одне з головних воріт до Мекки, яке мусульмани повинні відвідати принаймні раз у житті.} & \textbf{\textcolor[HTML]{DC3220}{Output}\ } $\blacktriangleright$ {\fontencoding{T2A}\selectfont It is one of the main gateways to Mecca that Muslims must visit at least once in their lifetime.}  \\
\addlinespace
\cdashline{1-2}[1pt/1pt]
\addlinespace


\textbf{\textcolor[HTML]{005AB5}{Coherence Input}\ }
 $\blacktriangleright$ {\fontencoding{T2A}\selectfont Виправте зв'язність в цьому тексті: Зайферт: Ця зміна здавалася певною протягом більшої частини року. Нещодавно Гуделл сказав, що очікує голосування під час зборів власників ліги в березні.
} & \textbf{\textcolor[HTML]{005AB5}{Coherence Input}\ }
 $\blacktriangleright$ {\fontencoding{T2A}\selectfont Correct the coherence in this text: Seifert: This change seemed certain for most of the year. Goodell recently said he expects a vote at the league's owners meeting in March.
} \\

\textbf{\textcolor[HTML]{DC3220}{Output}\ } $\blacktriangleright$ {\fontencoding{T2A}\selectfont Зайферт: Ця зміна здавалася певною протягом більшої частини року, але нещодавно Гуделл сказав, що очікує голосування під час зборів власників ліги в березні.
} & \textbf{\textcolor[HTML]{DC3220}{Output}\ } $\blacktriangleright$ {\fontencoding{T2A}\selectfont Seifert: That change seemed certain for most of the year, but Goodell recently said he expects a vote at the league's owners meeting in March.
} \\
\addlinespace
\cdashline{1-2}[1pt/1pt]
\addlinespace


\textbf{\textcolor[HTML]{005AB5}{Paraphrasing Input}\ }
 $\blacktriangleright$ {\fontencoding{T2A}\selectfont Перефразуйте це речення: Чоловік грає на музичній клавіатурі.}  & \textbf{\textcolor[HTML]{005AB5}{Paraphrasing Input}\ } $\blacktriangleright$ {\fontencoding{T2A}\selectfont Rephrase this sentence: The man is playing the musical keyboard.}\\
 
\textbf{\textcolor[HTML]{DC3220}{Output}\ } $\blacktriangleright$  {\fontencoding{T2A}\selectfont Чоловік грає на клавіатурі.} & \textbf{\textcolor[HTML]{DC3220}{Output}\ } $\blacktriangleright$  {\fontencoding{T2A}\selectfont A man plays the keyboard.}\\ 

\bottomrule
\end{tabular}
\caption{Example errors made by \textsc{\method-aya-101} model for all tasks with English translations.}
\label{tab:qualitative_outputs_errors}
\end{table*}

\subsection{Qualitative Error Analysis}
In this section, we first discuss the subpar performance of most baseline models on GEC, as observed in Table \ref{tab:summary_results}. Careful inspection of the model outputs indicates several problems with zero-shot model evaluation. The most frequent problems include repeated generation, output generation in English instead of Ukrainian, explanation of corrections made, text generation indicating no change is needed, to name a few. These models also suffer from an overcorrecting issue \cite{fang2023chatgpt} and tend to perform paraphrasing and fluency rewrites. As a result, in many cases, the conservative span-based $F_{0.5}$ metric (used to evaluate GEC) can't capture the correct edits, resulting in low performance. 

Next, we evaluate one of our best-performing models (\method-aya-101) qualitatively. For each task, we provide the model a sample input along with an instruction on what to do and show the model-generated output for a handful of such inputs in Table \ref{tab:qualitative_outputs}. We also highlight some of the errors made by our model in Table \ref{tab:qualitative_outputs_errors}. The English translations for all examples are provided in the same tables for reference. On the GEC task, the output quality outperforms all baseline models. Due to instruction tuning, the edits become more conservative and therefore, are better captured by $F_{0.5}$ metric using M2scorer\footnote{\url{https://github.com/nusnlp/m2scorer}}. The instruction-tuned models avoid common errors such as repetitions and generation of gibberish text and are much better at following instructions. However, the edits made are not always correct. For the simplification task, the majority of errors arise from changes in meaning due to excessive text truncation. Another typical negative pattern is the filtration of named entities and/or their replacement with pronouns. The coherence task is performed rather successfully. The model either edits the text correctly or leaves the text uncorrected. The most common issue is the incorrect usage of conjunctions, disrupting the logical flow, e.g. using ``but'' instead of ``and'', ``however'' instead of ``so'', etc. Paraphrasing is done mainly on the lexical level by changing the word or phrase order inside the text. In longer texts, such as those in the MRPC dataset, we sometimes observe a change in meaning compared to the input, whereas in shorter texts, such as those in STS and QQP, it tends to align more closely with the reference rewrites. Errors highlighting some of these problems are shown in Table \ref{tab:qualitative_outputs_errors}.

\section{Conclusions}
We introduce \methodsp -- an instruction-tuned LLM for Ukrainian text editing and corresponding \method-Instruct dataset. We describe in detail the construction of \method, including how we curate the instruction dataset in Ukrainian for text editing tasks. We empirically show that \methodsp significantly outperforms other models on text editing tasks. We also analyze the effect of modeling choices (scale and architecture) on task performance. Overall, our experiments support the hypothesis that domain and task-specific instruction tuning is needed to obtain better performance on complex text editing tasks. Finally, all our datasets and models are released to the community to help advance research in the area of Ukrainian NLP.

\section{Limitations}
\label{sec:limitations}
While we have introduced an instruction-tuned LLM for Ukrainian text editing, we acknowledge a few limitations of our work. First, due to the limitations of the translation API used, our training data may not be of the highest quality. This limitation could potentially be overcome by curating high-quality data from native speakers of the Ukrainian language. The scale of our training data can also be increased over our current set. 

Second, while we use standard evaluation metrics for the text editing tasks, we acknowledge that many of these metrics have limitations and do not capture many aspects of text quality (e.g. meaning preservation, etc.). For example, in the case of evaluating paraphrasing, there is no single automatic metric that holistically captures all important aspects of a good paraphrase as judged by humans. In fact, even if one narrows down to using BLEU score as a metric,  it has been shown in prior work that either one of reference-free BLEU score or reference-based BLEU score may correlate better with human judgments, and this may be dataset or benchmark-dependent \cite{shen2022evaluation} which is why we report both reference-based and reference-free BLEU scores in our evaluations for paraphrasing. In addition to BLEU, one would also report a semantic similarity score (like BERTScore) between the paraphrase and the source to capture how semantically close the paraphrase is to the source (or reference). For English, this is typically done using popular sentence embedding models like BERT, but it is not clear what the best approach is for Ukrainian, which is why we do not consider this dimension in our evaluation. One could potentially address such limitations by directly seeking human judgments on the quality of model predictions.

Finally, while we explore different settings of hyper-parameters (like batch size and learning rate) and different variants of prompts in our experiments, our search space is not exhaustive and is limited due to computational budgets and time constraints. We also acknowledge that the performance of closed models like GPT4 may drift or change over time due to model refreshes. Even in cases where model artifacts were publicly available, one must acknowledge that they were likely pre-trained on different datasets in the pretraining stage, and the precise effect of this on our specific downstream task performance is not known and is absorbed in our model performance reports. Research around an improved characterization of such variance in expected performance would be useful in the future.

\section{References}\label{sec:reference}

\bibliographystyle{lrec_natbib}
\bibliography{main_paper}

\label{lr:ref}
\bibliographystylelanguageresource{lrec-coling2024-natbib}

\clearpage
\section{Appendix A}





\begin{table}[h]
\small
\centering
\begin{tabular}{p{0.45\textwidth} p{0.45\textwidth}}
\toprule
\textbf{\textcolor[HTML]{005AB5}{GEC Input}\ }
 $\blacktriangleright$ {\fontencoding{T2A}\selectfont Виправте граматику в цьому реченнi: А ти, батюшка, стало бути, тут в сторожi?} &  \textbf{\textcolor[HTML]{005AB5}{GEC Input}\ } $\blacktriangleright$ {\fontencoding{T2A}\selectfont Correct the grammar in this sentence: And you, father, are you here in guard duty?}\\
  
 \textbf{\textcolor[HTML]{DC3220}{Output}\ } $\blacktriangleright$ {\fontencoding{T2A}\selectfont А ти, батюшко, стало бути, тут у сторожi?} &  \textbf{\textcolor[HTML]{DC3220}{Output}\ } $\blacktriangleright$ {\fontencoding{T2A}\selectfont And you, father, are you here on guard duty?} 
\\
\addlinespace
\cdashline{1-2}[1pt/1pt]
\addlinespace


\textbf{\textcolor[HTML]{005AB5}{Coherence Input}\ }
 $\blacktriangleright$ {\fontencoding{T2A}\selectfont Покращiть зв’язнiсть тексту: Тодi вiн вiдвiдав острови Грецiї. Хвороба змусила його повернутися до Риму.} & \textbf{\textcolor[HTML]{005AB5}{Coherence Input}\ }  $\blacktriangleright$ {\fontencoding{T2A}\selectfont Improve the coherence of the text: Then he visited the islands of Greece. Illness forced him to return to Rome.}\\

\textbf{\textcolor[HTML]{DC3220}{Output}\ } $\blacktriangleright$  {\fontencoding{T2A}\selectfont Потiм вiн вiдвiдав острови Грецiї, але хвороба змусила його повернутися до Риму.} & \textbf{\textcolor[HTML]{DC3220}{Output}\ } $\blacktriangleright$ {\fontencoding{T2A}\selectfont He then visited the islands of Greece, but illness forced him to return to Rome.} \\
\addlinespace
\cdashline{1-2}[1pt/1pt]
\addlinespace


\textbf{\textcolor[HTML]{005AB5}{Simplification Input}\ }
 $\blacktriangleright$ {\fontencoding{T2A}\selectfont Спростiть речення: вода спочатку холодна, а потiм волога.} & \textbf{\textcolor[HTML]{005AB5}{Simplification Input}\ }
 $\blacktriangleright$ {\fontencoding{T2A}\selectfont Simplify the sentence: first the water is cold, and then it is wet.}  \\
\textbf{\textcolor[HTML]{DC3220}{Output}\ } $\blacktriangleright$ {\fontencoding{T2A}\selectfont Вода i холодна, i мокра.} & \textbf{\textcolor[HTML]{DC3220}{Output}\ } $\blacktriangleright$ {\fontencoding{T2A}\selectfont The water is both cold and wet.}  \\
\addlinespace
\cdashline{1-2}[1pt/1pt]
\addlinespace


\textbf{\textcolor[HTML]{005AB5}{Paraphrasing Input}\ }
 $\blacktriangleright$ {\fontencoding{T2A}\selectfont Перефразуйте речення: Наскiльки я можу судити, обидва ще живi.}  & \textbf{\textcolor[HTML]{005AB5}{Paraphrasing Input}\ }
 $\blacktriangleright$ {\fontencoding{T2A}\selectfont Rephrase the sentence: As far as I can tell, both are still alive.} \\
 
\textbf{\textcolor[HTML]{DC3220}{Output}\ } $\blacktriangleright$ {\fontencoding{T2A}\selectfont Наскiльки я розумію, вони обидва ще живi.} & \textbf{\textcolor[HTML]{DC3220}{Output}\ } $\blacktriangleright$ {\fontencoding{T2A}\selectfont As far as I understand, they are both still alive.}\\ 

\bottomrule
\end{tabular}
\begin{minipage}[t]{\textwidth}
\hspace{-2in}
\caption{Example model inputs and outputs of the text editing tasks that \methodsp can perform. \\English translations of the examples in Figure \ref{fig:main_examples} are  provided for reference.\newline}
\end{minipage}
\label{tab:fig_1_translations}
\end{table}

\begin{table}[htb!]
\small
\centering
\begin{tabular}{@{}l|c|c|c|c|c|c|c@{}}
\toprule
\textbf{Model} & \multicolumn{7}{c}{\textbf{Text Editing Tasks}} \\ 
\toprule
 & \multicolumn{2}{c|}{\textbf{Simplification}} & \multicolumn{2}{c|}{\textbf{Coherence}} & \multicolumn{3}{c}{\textbf{Paraphrasing}} \\
\cmidrule{2-8} 
 & Asset & Turk & Sports & Wiki & MRPC & STS & QQP \\
\toprule
Copy & 17.75 & 24.04 & 26.61 & 28.37 & 100/39.90 & 100/38.80 & 100/26.20 \\ 
\toprule
\textsc{Bactrian-X-7b} & 36.02 & 37.13 & 40.7 & 38.62 & 65.5/29.20 & 45.6/20.4 & 13.5/4 \\
\textsc{UAlpaca-7b} & 33.54 & 35.96 & 32.48 & 33.45 & 57.6/24.2 & 20.5/9.6 & 6.2/1.8 \\
\textsc{Mistral-7b} & 39.85 & 38.54 & 32.75 & 30.58 & 37.6/18 & 16.9/8.3 & 5.9/2 \\ 
\textsc{mt0-large} & 32.91 & 27.94 & 22.32 & 21.20 & 10.8/5.2 & 4.7/2.3 & 4.7/1.3 \\
\textsc{aya-101} & 32.02 & 37.32 & 38.42 & 37.68 & 78.5/34.5 & 56.8/25 & 32.1/9.8 \\
\toprule
\textsc{GPT-3.5-Turbo} & 42.52 & 39.04 & 44.84 & 45.44 & 33.2/17.6 & 24/12.4 & 22.9/9 \\
\textsc{GPT4} & 42.20 & 39.05 & 43.35 & 43.96 & 29.2/16.1 & 17/12.7 & 19.9/9 \\ 
\toprule
\textsc{\method-Bactrian-X-7b} & 35.15 & 37.75 & 47.29 & 50.53 & 63.2/29.1 & 67.1/31.9 & 66.5/20.2 \\
\textsc{\method-Mistral-7b} & 31.73 & 35.92 & 44.01 & 44.72 & 75.1/31.8 & 81/31.7 & 77.3/21.3 \\
\textsc{\method-Llama-2-7b-chat} & 39.29 & 35.80 & 48.13 & 51.95 & 46.2/22.3 & 50.7/22.3 & 50.5/16.7 \\
\textsc{\method-Llama-2-13b-chat} & 37.09 & 36.93 & 47.54 & 53.94 & 55.5/26.6 & 57.9/24.6 & 58.3/18.1 \\
\textsc{\method-mt5-large} & 34.82 & 37.17 & 47.97 & 49.87 & 71/32 & 78/34.8 & 80.7/23.3 \\
\textsc{\method-mt0-large} & 33.85 & 37.28 & 48.25 & 48.41 & 71.5/32.3 & 79.4/34.3 & 81.2/23.2 \\
\textsc{\method-mt5-xxl} & 38.50 & 37.52 & 48.87 & 49.53 & 67.3/30.9 & 69.1/30.5 & 75.3/22.3 \\
\textsc{\method-mt0-xxl-mt} & 38.95 & 38.20 & 48.67 & 53.80 & 65.4/30.4 & 69.6/34.8 & 70.4/21.6 \\
\textsc{\method-aya-101} & 37.71 & 37.95 & 47.87 & 51.94 & 69.9/31.6 & 71.7/33.3 & 74.2/22.5 \\
\bottomrule
\end{tabular}
\begin{minipage}[t]{\textwidth}
\hspace{-2in}
\caption{Comparison of \methodsp models against various baselines, categorized by constituent datasets. We report detailed metrics for each dataset within a task. GEC is not relevant here since it is a single dataset. For Simplification and Coherence, we report SARI. For Paraphrasing, we report reference-free / reference-based BLEU just as in Table \ref{tab:summary_results}. All scores have been scaled to lie between 0 and 100.}
\end{minipage}
\label{tab:full_results_per_data_source}
\end{table}


\begin{table*}[h!]
\footnotesize
\centering
\begin{tabular}{p{0.13\textwidth}p{0.39\textwidth}p{0.39\textwidth}}
\toprule
\textbf{Task} & \textbf{Verbalizers} & \textbf{English translation}\\
\toprule
\textbf{GEC} & \begin{tabular}[c]{@{}l@{}}{\fontencoding{T2A}\selectfont``Виправте граматику в цьому реченні:''}\\ {\fontencoding{T2A}\selectfont``Виправте граматичні помилки в цьому} \\ 
{\fontencoding{T2A}\selectfont реченні:''} \\ 
{\fontencoding{T2A}\selectfont``Удосконаліть граматику цього тексту:''}\\
{\fontencoding{T2A}\selectfont``Виправте всі граматичні помилки:''}\\
{\fontencoding{T2A}\selectfont``Зробіть речення граматичним:''}\\
{\fontencoding{T2A}\selectfont``Видаліть граматичні помилки:''}\\
{\fontencoding{T2A}\selectfont``Виправте помилки в цьому тексті:''}\\
{\fontencoding{T2A}\selectfont``Виправте граматичні помилки:''}\\
{\fontencoding{T2A}\selectfont``Виправити граматику:''}\end{tabular} 
& 
\begin{tabular}[c]{@{}l@{}}``Correct the grammar in this sentence:''\\ ``Correct the grammatical errors in this\\
sentence:''\\ 
``Improve the grammar of this text:''\\
``Correct all grammatical errors:''\\
``Make the sentence grammatical:''\\
``Remove grammatical errors:''\\
``Correct the errors in this text:''\\
``Correct the grammatical errors:''\\
``Correct the grammar:''\end{tabular}
\\ \hline 
\textbf{Simplification} & \begin{tabular}[c]{@{}l@{}}{\fontencoding{T2A}\selectfont``Спростіть речення:''}\\
{\fontencoding{T2A}\selectfont``Напишіть простішу версію для речення:''}\\

{\fontencoding{T2A}\selectfont``Спростіть це речення:''}\\
{\fontencoding{T2A}\selectfont``Зробіть речення простим:''}\\
{\fontencoding{T2A}\selectfont``Спростіть цей текст:''}\\
{\fontencoding{T2A}\selectfont``Перепишіть речення так, щоб воно було}\\ {\fontencoding{T2A}\selectfont простішим:''}\\
{\fontencoding{T2A}\selectfont``Перепишіть це речення простіше:''}\\
{\fontencoding{T2A}\selectfont``Зробіть речення простіше:''}\\
{\fontencoding{T2A}\selectfont``Спростіть цей текст:''}\\
{\fontencoding{T2A}\selectfont``Використовуйте простіші слова:''}\\
{\fontencoding{T2A}\selectfont``Зробіть цей текст легше для розуміння:''}\end{tabular} 
& \begin{tabular}[c]{@{}l@{}}
``Simplify the sentences:''\\
``Write a simpler version for the sentence:''\\
``Simplify this sentence:''\\
``Make the sentence simple:''\\
``Simplify this text:''\\
``Rewrite the sentence so that it is simpler:''\\
\\
``Rewrite this sentence more simply:''\\
``Make the sentences simpler:''\\
``Simplify this text:''\\
``Use simpler words:''\\
``Make this text easier to understand:''\end{tabular}
\\ \hline 
\textbf{Coherence} & \begin{tabular}[c]{@{}l@{}}{\fontencoding{T2A}\selectfont``Виправте зв'язність в реченні:''}\\
{\fontencoding{T2A}\selectfont``Покращіть зв'язність тексту:''} \\
{\fontencoding{T2A}\selectfont``Виправте зв'язність в цьому тексті:''}\\
{\fontencoding{T2A}\selectfont``Виправте відсутність зв'язності в реченні:''}\\
{\fontencoding{T2A}\selectfont``Виправте зв'язність в тексті:''}\\
{\fontencoding{T2A}\selectfont``Виправте зв'язність речення:''}\\
{\fontencoding{T2A}\selectfont``Зробіть текст більш зв'язним:''}
\end{tabular} 
& 
\begin{tabular}[c]{@{}l@{}}
``Correct the coherence in the sentence:''\\
``Improve text coherence:'' \\
``Correct the coherence in this text.''\\
``Correct the lack of coherence in the sentence:''\\
``Correct the coherence in the text:''\\
``Correct the coherence of the sentence:''\\
``Make the text more coherent:''
\end{tabular}
\\ \hline
\textbf{Paraphrasing} & \begin{tabular}[c]{@{}l@{}}{\fontencoding{T2A}\selectfont``Перефразуйте речення:''}\\
{\fontencoding{T2A}\selectfont``Перепишіть речення іншими словами:''}\\
{\fontencoding{T2A}\selectfont``Перефразуйте цей текст:''}\\
{\fontencoding{T2A}\selectfont``Перефразуйте це речення:''}\\
{\fontencoding{T2A}\selectfont``Перефразуйте:''}\\
{\fontencoding{T2A}\selectfont``Напишіть перефраз для речення:''}\\
{\fontencoding{T2A}\selectfont``Напишіть перефразовану версію речення:''}\\
{\fontencoding{T2A}\selectfont``Перепишіть це речення:''}\\
{\fontencoding{T2A}\selectfont``Перепишіть цей текст:''}\\
{\fontencoding{T2A}\selectfont``Переформулюйте це речення:''}\\
{\fontencoding{T2A}\selectfont``Перефразуйте це речення:''}\\
{\fontencoding{T2A}\selectfont``Переформулюйте цей текст:''}\end{tabular} 
& \begin{tabular}[c]{@{}l@{}}
``Rephrase the sentence:''\\
``Rewrite the sentence in other words:''\\
``Paraphrase this text:''\\
``Rephrase this sentence:''\\
``Paraphrase:''\\
``Write a paraphrase for the sentence:''\\
``Write a paraphrased version of the sentence:''\\
``Rewrite this sentence:''\\
``Rewrite this text:''\\
``Rephrase this sentence:''\\
``Paraphrase this sentence.''\\
``Rephrase this text:'' \end{tabular}
\\ \hline
\end{tabular}
\caption{A complete list of verbalizers for each task used as instructions in the \method-Instruct dataset. The English translations are provided for reference.}
\label{tab:verbalizers_full}
\end{table*}

\end{document}